\documentclass[conference]{IEEEtran}
\IEEEoverridecommandlockouts
\usepackage{cite}
\usepackage{amsmath,amssymb,amsfonts}
\usepackage{algorithmic}
\usepackage{graphicx}
\usepackage{textcomp}
\usepackage{xcolor}
\def\BibTeX{{\rm B\kern-.05em{\sc i\kern-.025em b}\kern-.08em
    T\kern-.1667em\lower.7ex\hbox{E}\kern-.125emX}}
    
\usepackage{hyperref}

\begin{document}

\title{Video-based cattle identification and action recognition
\thanks{This research was funded by 
 \href{https://sief.org.au}{Science and Industry Endowment Fund} }
}

\author{
\IEEEauthorblockN{Chuong Nguyen, Dadong Wang}
\IEEEauthorblockA{\textit{Imaging and Computer Vision} \\
\textit{CSIRO Data61}\\
Australia \\
\{chuong.nguyen, dadong.wang\}@csiro.au}
\and
\IEEEauthorblockN{Karl Von Richter, Philip Valencia}
\IEEEauthorblockA{\textit{Robotic and Autonomous Systems} \\
\textit{CSIRO Data61}\\
Australia \\
\{karl.vonrichter, philip.valencia\}@csiro.au}
\and
\IEEEauthorblockN{Flavio A. P. Alvarenga}
\IEEEauthorblockA{\textit{Health and Resilience} \\
NSW Department of Primary Industries, Livestock Industry Centre, UNE\\
Armidale, NSW 2351, Australia\\
\textit{CSIRO Agriculture and Food}\\
FD McMaster Laboratory Chiswick, Armidale, NSW 2350, Australia \\
flavio.alvarenga@dpi.nsw.gov.au}
\and
\IEEEauthorblockN{Gregory Bishop-Hurley}
\IEEEauthorblockA{\textit{Health and Resilience} \\
\textit{CSIRO Agriculture and Food}\\
Australia \\
greg.bishop-hurley@csiro.au}
}

\maketitle

\begin{abstract}
We demonstrate a working prototype for the monitoring of cow welfare by automatically analysing the animal behaviours. Deep learning models have been developed and tested with videos acquired in a farm, and a precision of 81.2\% has been achieved for cow identification. An accuracy of 84.4\% has been achieved for the detection of drinking events, and 94.4\% for the detection of grazing events. Experimental results show that the proposed deep learning method can be used to identify the  behaviours of individual animals to enable automated farm provenance. Our raw and ground-truth dataset will be released as the first public video dataset for cow identification and action recognition. Recommendations for further development are also provided.

\end{abstract}


\section{Introduction}
Australia has a reputation for safe and high-quality food. Trust in the provenance of that food - including region of origin, sustainability and ethical production - creates premium products that command higher prices and open new markets for Australian producers. Most of the provenance claims valued by consumers relate to the farm where the animals are raised. These include origin (region or country), animal welfare and sustainable practices. As part of automated farm provenance, we have conducted experiments and research to automatically detect and classify animal behaviours. 

\begin{figure}[!t]
\centering
\includegraphics[width=\linewidth]{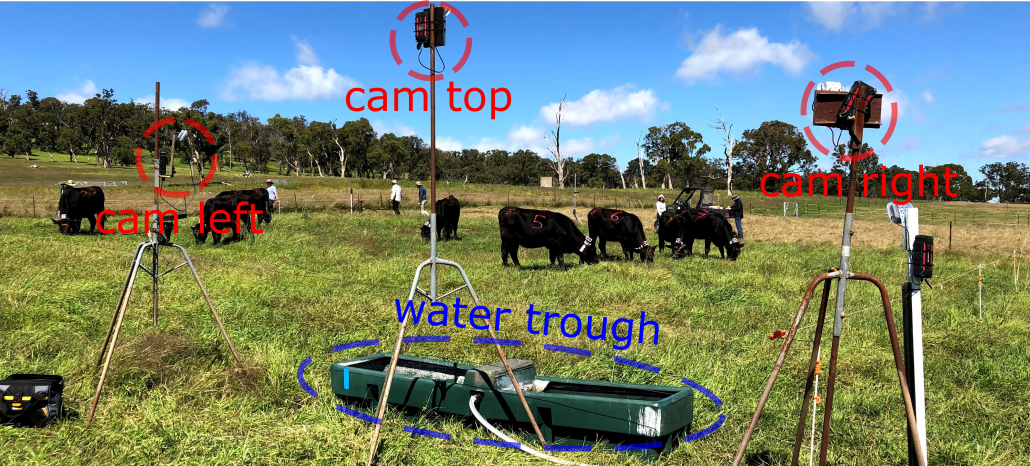}
\caption{Experimental setup for cow video-based identification and behaviour classification.}
\label{fig:exp_setup}
\end{figure}

This research component focuses on monitoring cattle’s activities and behaviours, particularly drinking events, using data annotation from video recordings of cattle near the water trough area as shown in Fig. \ref{fig:exp_setup}. 
This information was then used to label signals from motion sensors on cattle's collars and ear tags to train embedded machine-learning algorithms. As manual labelling of cattle behaviours from videos (and synchronising with motion sensor signals) could be significantly labour-intensive \cite{borchers2016validation}, there is a need for an image-based machine learning solution to automate this process once there is enough labelled image data.
The problems this project aims to solve include:
\begin{itemize}
    \item What camera views are optimal for capturing informative video streams of cow behaviours, particularly drinking behaviour events?
    \item What is the best way to annotate video?
    \item How to identify individual cows from video?
    \item How to classify cow behaviours from video and associate behaviours with cow IDs?
\end{itemize}

A related work \cite{jingqiu2017cow} proposes bounding-box cattle detection and single-frame behaviour recognition for hoof disease and estrous behaviour based on the body curvature and overlapping between bounding boxes. However this work does not identify individual cows and their subtle behaviours such as drinking versus grazing, so it is difficult to know which cow is doing what. Existing approaches to identify individual cows includes using nose patterns \cite{bello2020cattle} or body patterns \cite{bello2020image} or rear-view videos \cite{qiao2019individual}, however these require a well-control environment and/or close-up image capture that are difficult to set up in the field. To reduce the complexity and cost of cattle behaviour recognition, motion sensors such as accelerometers and gyroscopes associated with cattle IDs \cite{rahman2018cattle} have been used but these lack behaviour annotation. Therefore video recording and labelling are required to generate annotations for machine learning algorithms to recognise behaviours from sensor signals \cite{wang2018development, tian2021real}.

In this paper, we propose a solution to detect, identify and recognise behaviours of individual cattle. We collected videos from multiple cameras, annotated video for identifying individual cattle and their action, and then developed an image-based deep-learning solution to automate the process. 
Source code for a baseline solution, raw videos and data generated from this project are released at \url{https://github.com/chuong/cattle_identification_action_recognition}.

\section{Methodology}

There are a few challenges when working with cows on the field:
\begin{itemize}
    \item It is challenging to identify similar animals from a larger herd on the field. Because the cows are all black and have similar size in our experiment, cow identification based on body patterns \cite{bello2020image} or nose print \cite{bello2020cattle} cannot be used. Furthermore, cow identification based on rearview and body movements \cite{qiao2019individual} cannot be applied to uncontrolled environment on the field. This in turn makes it difficult to associate their activities with their ID. 
    \item Behaviour annotation is challenging, particularly for non-expert humans, to distinguish as many behaviours look similar. Furthermore, the transition between different behaviours is sometimes not clear-cut.
    \item It is difficult to synchronise the video recordings while recording asynchronously on the field. 
    \item Videos are long and require significant resources to generate a complete set of annotations. 
    \item There is significant self occlusion when the animals get close together. The action recognition could be view dependent.
\end{itemize}

As these are challenging tasks to humans, they are also challenging to train a computer to perform the same task. As a result, a number of technical solutions are used to get around these problems. The experiment was approved by the CSIRO FD McMaster Laboratory Chiswick Animal Ethics Committee with the animal research authority number 17/20.

\subsection{Data collection }

Multiple cameras fisheye cameras (GoPro 5 Black) with built-in GPS receiver were installed at CSIRO Armidale site to record videos. GPS signal embedded in the recorded video provided accurate timestamps to synchronise different cameras within the uncertainty of a frame. External battery power banks were used to power the cameras throughout the day under any weather conditions. 

\begin{figure}[!t]
\centering
\includegraphics[width=\linewidth]{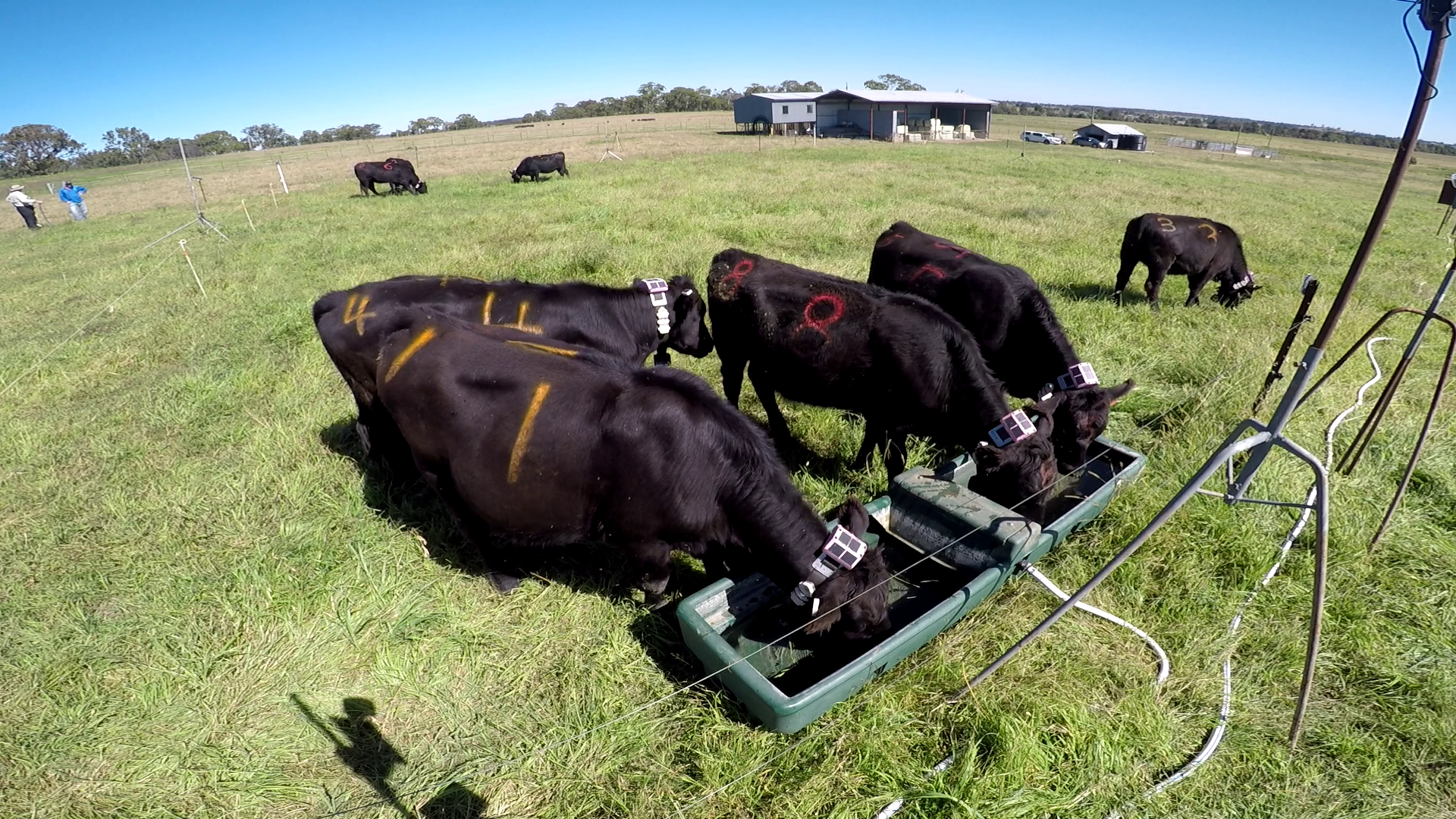}
\includegraphics[width=\linewidth]{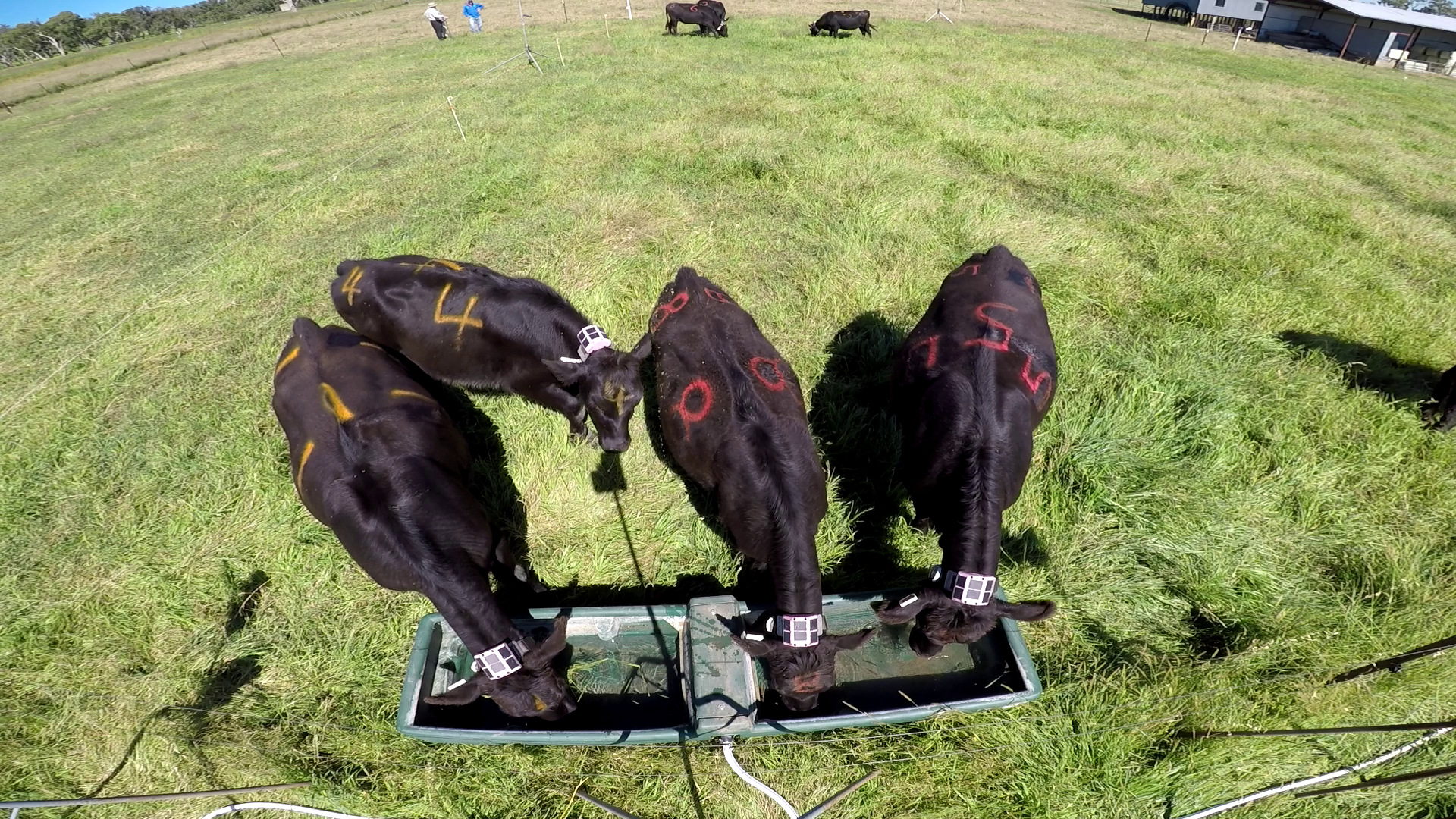}
\includegraphics[width=\linewidth]{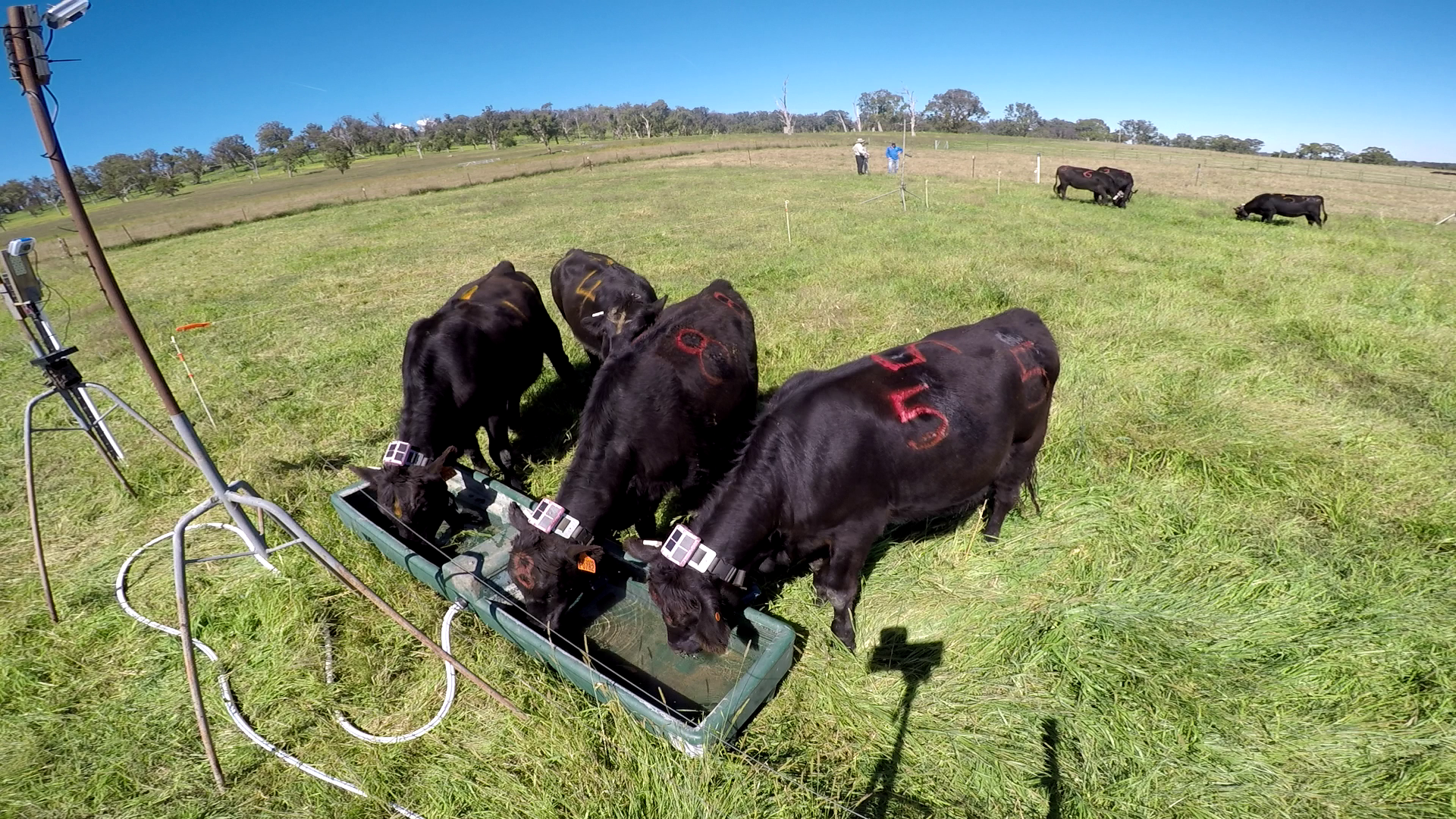}
\caption{Left, top and right views from water trough cameras.}
\label{fig:water_trough_cams}
\end{figure}

To tell the cows apart, numbers were handwritten on different parts of the cow body so that at least one of these are visible from any view angle and close enough distances as shown in Fig. \ref{fig:water_trough_cams}. Fortunately, the different shapes of the limited list of numbers are good enough for a deep network to distinguish without explicit handwriting recognition.

\subsection{Data annotation}
Due to long video recordings, annotation is only applied when at least one cow comes close to the main camera. Particularly, this work focuses on drinking action, all video segments containing drinking actions occurring at the water trough are marked from the videos recorded by the top camera. Then the duration of each video segment is doubled and centred on the original segment before the annotation is applied to the frames within the doubled duration. This allows to include other non-drinking behaviour to be observed and annotated without causing a significant class-imbalanced problem.

Only key frames are selected for ID annotation to save the annotation effort. Linear interpolation between the key frames gives good enough accuracy.
Cow IDs and associated bounding boxes are manually annotated on the key frames and saved as a ground truth dataset for identification. 
The cow ID and bounding box annotations are visualised to verify the accuracy.

Behaviour annotation is performed separately and directly on videos and stored in VTT subtitle \cite{W3C2020} files. This includes cow ID, the start \& end time points of each behaviour or action. VLC player is used to play the video and verify the annotation from the subtitles.

\begin{figure}[!t]
\centering
\includegraphics[width=\linewidth]{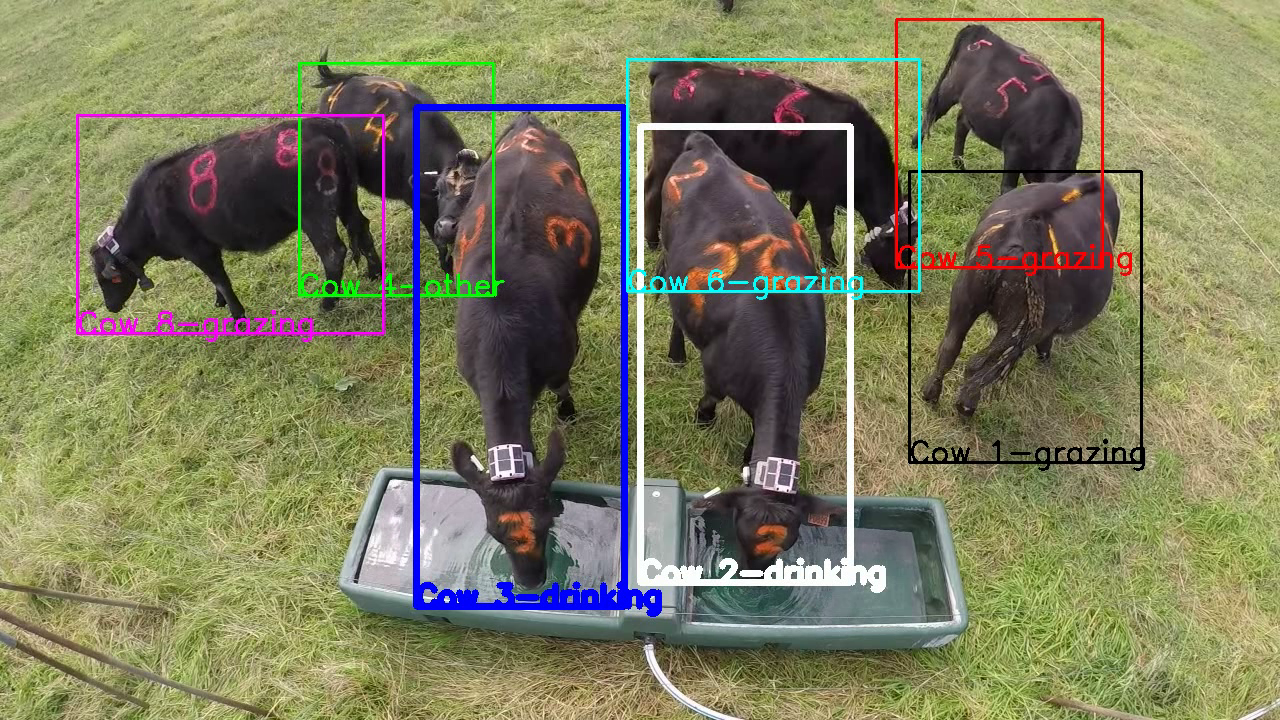}
\caption{Annotation of cow IDs and behaviours.}
\label{fig:cow_annotations}
\end{figure}

Both cow ID and behaviour annotation is shown together in Fig. \ref{fig:cow_annotations}.

\subsection{Approaches in object detection}
An object detection network is needed for cow ID detection. This could be done using popular deep network architectures such as YOLO \cite{redmon2016you} and Faster R-CNN \cite{ren2015faster}. Unlike conventional object detection and classification where objects have distinct appearance, cows in this experiment look almost identical except the written number on their bodies. As a result, the cow detection is expected to work well, but cow classification for identification is challenging. As a result, Cascade R-CNN \cite{cai2018cascade} was chosen as it is expected to be better at fine-grained object detection and classification.

\subsection{Approaches in action recognition}
There are two approaches for action detection for cow behaviours. The first approach is detection of single action of a cow identified by previous ID detection. This approach requires a video containing only a single cow in the middle of the image. The second approach is action recognition for multiple objects in an image. This approach includes both object detection and action recognition.

Unfortunately, existing network architectures for the second approach such as YOWO \cite{mo2020towards} and MOC-Detector \cite{li2020actions} require training from scratch on new datasets and do not train well on our cow dataset. One reason is the limited size of the dataset. Another reason is that the cow actions look very similar to each other, and their heads point toward the ground most of the time and make their actions look very similar. This makes it difficult for the second action recognition approach to distinguish similar cows and recognise their actions at the same time. As a result, we have to rely on the first approach. One candidate for action recognition is Temporal Segment Networks (TSN) \cite{wang2016temporal} which recognises action from a stack of frames, with optional input of optical flow between consecutive frames.

\section{Results}

\subsection{Data collection and preprocessing}
An experiment was carried out 
from 18/03/2020 to 27/03/2020. Every day, 
eight cows with painted IDs on their sides and heads to distinguish them were released into the monitoring area. 
A water trough with automatic refilling mechanism was installed on the side of the patch and connected to a water supply. 
The experiment protocol was approved by the Animal Research Authority.

Three GoPro Hero 5 Black \cite{GoProHero5Black2020} cameras were mounted on poles next to the water trough to focus on drinking behaviour events. 
The cameras were powered by large power banks connected via USB-C cables. 

Videos were recorded from the 3 cameras for 8 hours a day for 8 days.
The views from three cameras next to the water trough are shown in Fig. \ref{fig:water_trough_cams}.

The middle camera view shows clear separation between cows. As a result, video annotation was carried out on the videos recorded from this camera. 
 GPS data embedded in GoPro videos files could be extracted to obtain time stamp using FFMPEG \cite{Bellard2020} and GoPro-Utils \cite{Stillman2020, Irache2020}. Commands \texttt{ffmpeg} and \texttt{gpmd2csv} are used to extract GPS timestamp and location from GoPro videos.

Annotation for cow identification was performed using CVAT \cite{OpenVINO2020} annotation tool which could export to different data formats for deep learning. This annotation was carried out on the videos recorded for the 8 days of the experiment and exported to COCO dataset format. Training, validation and testing data are split in 0.70:0.05:0.25 ratio.

Annotation for cow behaviour was performed using VLC video player \cite{VideoLAN2020} and VTT subtitle \cite{W3C2020}. Cow ID and behaviour annotations are processed using a custom Python script to export to 
KINECTICS \cite{carreira2018short} dataset formats. 


Each cow is annotated with a bounding box and an ID using CVAT \cite{OpenVINO2020}. The annotation was performed on key frames, 1 for every 9 frames from the video. Linear interpolation is performed to obtain annotation for other frames between the key frames. The behaviour is recorded as VTT subtitle \cite{W3C2020} files which consist of time duration to the accuracy of second, for example:
\small{
\begin{verbatim}
0:05:11.000 --> 0:05:23.000
Cow 2 Drinking
0:05:17.000 --> 0:05:42.000
Cow 4 Other
0:05:22.000 --> 0:05:40.000
Cow 8 Grazing
\end{verbatim}
}

A visualisation of the combined ID, bounding box and behaviour annotations is shown in Fig. \ref{fig:cow_annotations}.
The annotations were exported to different formats including COCO \cite{lin2014microsoft} and KINECTICS \cite{carreira2018short} depending on the required input of deep learning platform. 

\begin{figure}[!t]
\centering
\includegraphics[width=\linewidth]{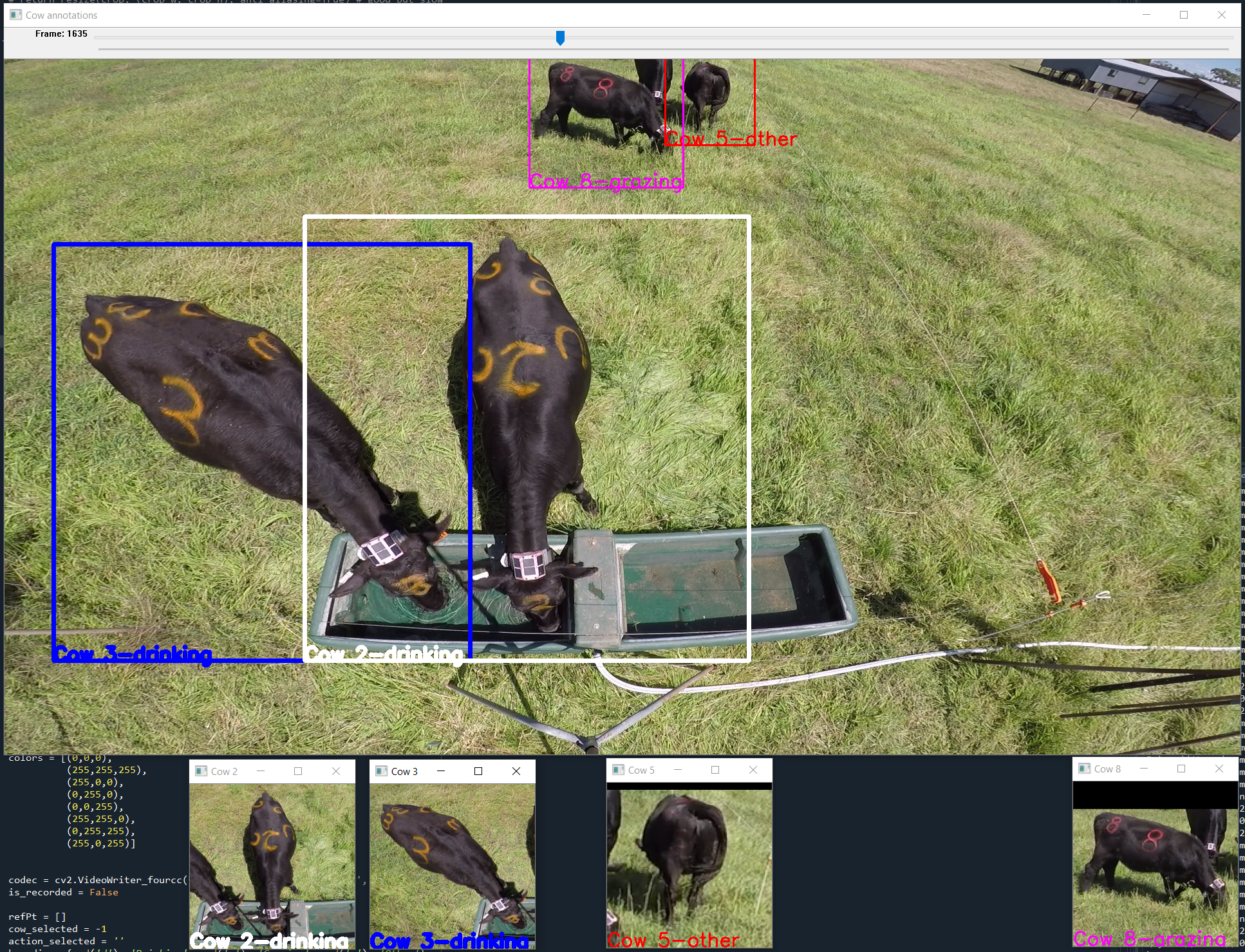}
\caption{Creating ground truth videos for behaviour recognition. Rectangular bounding boxes are converted by squared bounding boxes which are then resized to 256$\times$256 pixels as shown in small windows on the bottom of the screenshot.}
\label{fig:video_cropping}
\end{figure}

For behaviour recognition, video clips of each action are extracted to KINECTICS400 format with frame size of 256$\times$256 pixels. This is achieved by converting cow rectangular bounding boxes squared shape and scaled to this image size as shown in Fig. \ref{fig:video_cropping}. Additional zero padding is added for regions outside the input image so that the cow is mostly in the middle of the frame.

 A new cow behaviour dataset containing a single cow in each video frame following the format of Kinectics400 has been created. A total of 1715 videos were exported where "Drinking" class has 360 video, "Grazing" 413 videos and "Other" 942 video. Again training, validation and testing data are split in 0.70:0.05:0.25 ratio. 
 
\subsection{Cow identification}
\begin{figure}[!t]
\centering
\includegraphics[width=\linewidth]{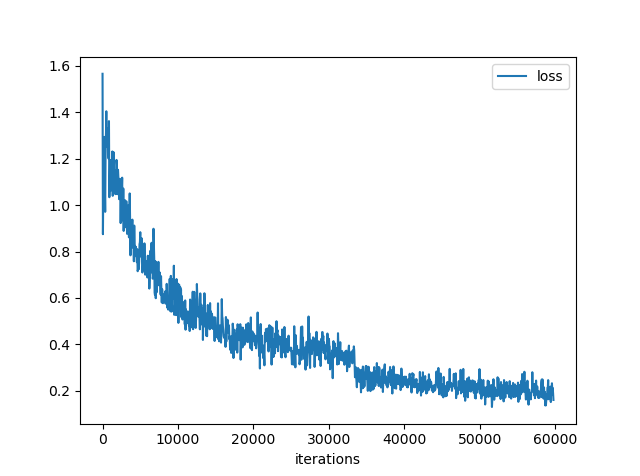}
\caption{Learning loss of Cascade RCNN training with starting learning rate of 0.01 and drop to 0.001. The training was performed for 7 epochs and stopped when the loss became plateau.}
\label{fig:learning_loss}
\end{figure}

Among different deep learning frameworks, we chose MMDetection framework \cite{Open-MMLab2020} which is based on PyTorch and supports multiple object classification and detection architectures and different data format. Cascade R-CNN shown in Fig \ref{fig:learning_loss} was trained with cow ID (8 classes) and bounding box annotation as ground truth data.
After training Cascade R-CNN on the cow's COCO dataset, the testing yielded average precision of 0.812 and average recall of 0.844. For comparison, the testing accuracy achieved by \cite{bello2020image} is 89.95\% on their private dataset which consists of cropped images captured from a relatively fixed view from a cow, while our cow images were captured at an arbitrary view.

\subsection{Cow behaviour classification}
We chose Temporal Segment Networks (TSN) \cite{wang2016temporal} without temporal information from MMAction2 library \cite{Open-MMLab2020} to train on KINECTICS \cite{carreira2018short} data export from our annotation. 
Finetuning a TSN action recognition model (pretrained on Kinectics400) over 100 epochs with learning rate of 8e-5 and decay of 1e-4 leads to an average accuracy of 0.72. More details of testing is shown in Table \ref{tab:TSN_accuracy}. The accuracy of "Drinking" and "Grazing" actions are quite high, while the accuracy of "Other" action is quite low. This is due to the fact that the "Other" videos contains some "Drinking" and "Grazing" actions that were missed out during annotation process.

\begin{table}[htbp]
\caption{Testing accuracy of Temporal Segment Network}
\begin{center}
\begin{tabular}{|c|c|c|c|c|c|}
\hline
& \textbf{Drinking} & \textbf{Grazing} & \textbf{Other} & \textbf{No. videos} & \textbf{Accuracy} \\
\hline
\textbf{Drinking} & 92 & 6 & 11 & 109 & 84.4\% \\
\hline
\textbf{Grazing} & 2 & 117 & 5 & 124 & 94.4\% \\
\hline
\textbf{Other} & 12 & 55 & 50 & 117 & 42.7\% \\
\hline
\end{tabular}
\label{tab:TSN_accuracy}
\end{center}
\end{table}

\subsection{Joint cow identification and behaviour recognition}
\begin{figure}[!t]
\centering
\includegraphics[width=\linewidth]{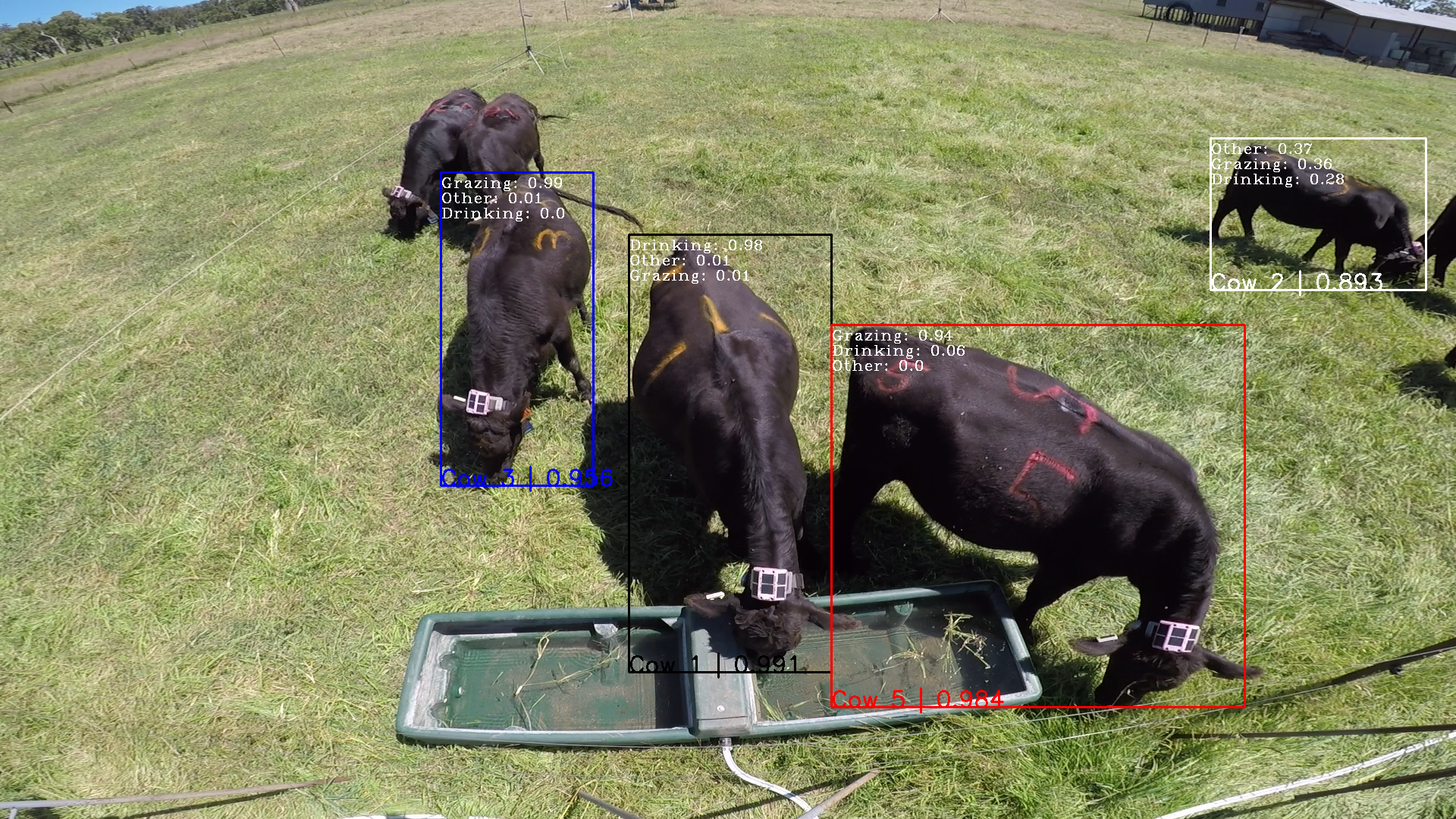}
\caption{An example of the output of the processing for cow ID and behaviour classification.}
\label{fig:cow_id_behaviour_recognition}
\end{figure}

To process a video stream to obtain cow ID and their behaviour, we combine the two networks into a single pipeline as shown in Fig. \ref{fig:cow_id_behaviour_recognition}. The first network takes a video frame and detect all cows and their bounding boxes. The second network takes a 1-second cropped video based on the bounding boxes of each cow and classifies its behaviour. The output of the pipeline is cow ID and the confidence of each behaviour as shown in Fig. \ref{fig:cow_id_behaviour_recognition}.

\begin{figure}[!t]
\centering
\includegraphics[width=\linewidth]{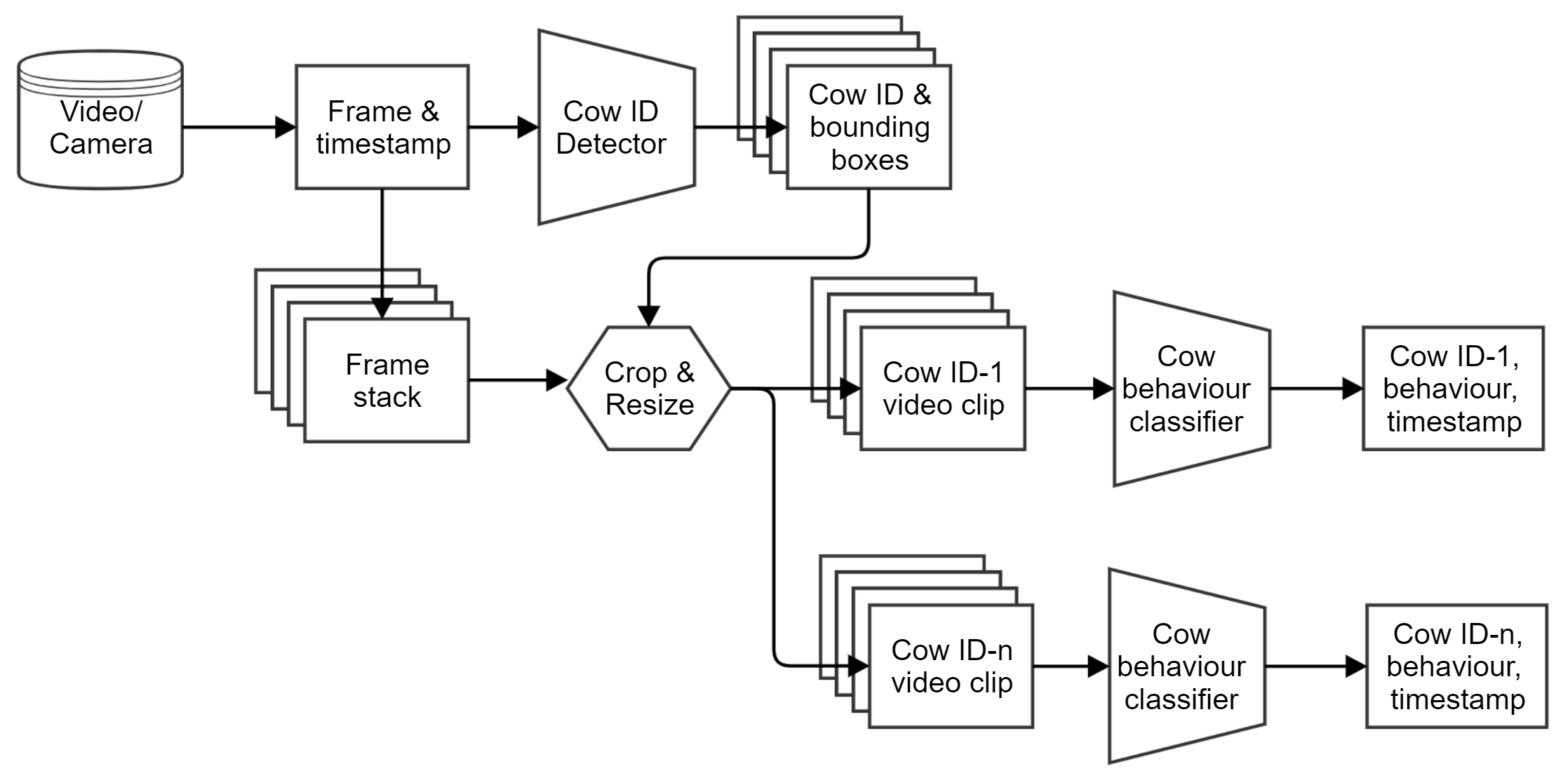}
\caption{Processing pipeline for cow ID and behaviour identification.}
\label{fig:cow_ID_behavious_iden_pipeline}
\end{figure}

\section{Discussion and future works}
This research has led to three important outcomes. First, a camera setup that could be used not only to acquire videos to annotate motion sensor data, but also to directly monitor cow welfare such as shy feeding. Second, annotated video data of cow IDs, bounding boxes and behaviours as ground truth for machine training based on the videos and other synchronised sensor data from collars and ear tags. Finally, a processing pipeline with two trained cow ID detection and behaviour classifiers that can automatically label new video streams captured from similar views.

We demonstrated a working prototype that can automatically monitor cow welfare on farm using low-powered embedded sensors and edge machine learning. This system will play a major role in evident-based quality control and automatic compliance in supply chain. 

There are several improvements for this system to be made for the deployment in production and large-scale environments:
\begin{itemize}
    \item Large-scale cattle identification and tracking. Our work is limited to a fixed and limited number of cattle (8) that are labelled and used to train a network. In reality, the number of cattle could be 100s or 1000s, so the current approach of using handwritten numbers to distinguish cows may not be practical. A possible approach is zero-shot identification and/or continuous 2D or 3D tracking. Additional sensors could be utilised to re-identify cattle.
    \item Multiview cattle identification and behaviour recognition. Although videos from 3 or more cameras were captured, only videos from Camera 1 were labelled and used in this study. Single view identification is not sufficient when there are more cattle and occlusion is unavoidable. Multiview camera approach will solve this problem and potentially improve the accuracy.
    \item Big data management and visualisation. The amount of data generated from multiple cameras is large and difficult to manage and synchronise. There is a need for a dedicated management system and a user interface to allow a user to navigate the data at any timestamp and across all available videos. This will also facilitate multiview data annotation and quality control.
\end{itemize}

\section*{Acknowledgment}
This research was funded by Science and Industry Endowment Fund \url{https://sief.org.au}. We acknowledge the following technical staff who have contributed to the research at CSIRO FD McMaster Laboratory Chiswick: Katie Austin (Technical Officer), Alistair Donaldson (Technical Officer) and Reg Woodgate (Senior Technical Officer) with NSW Department of Primary Industries, and Jody McNally (Research Technician) and Troy Kalinowski (Technical Officer) with CSIRO Agriculture and Food. Thanks are also due to Dr Ron Li and Mr Wyman Huang from Imaging and Computer Vision Group of CSIRO Data61 for their support.

\bibliographystyle{IEEEtran}
\bibliography{IEEEabrv,my_references}

\end{document}